\documentclass[12pt]{amsart}
\usepackage{ifthen,verbatim}
\usepackage{floatflt,graphicx,graphics}
\usepackage{caption}
\usepackage{subcaption}
\usepackage{multirow}
\usepackage{url}

 \usepackage{tikz} 
\usepackage{mathrsfs}
\usepackage{amsmath,amssymb,amsthm}
\usepackage{mathtools}

\numberwithin{equation}{section}
\nonstopmode
{\qed\bigskip}

\newcounter{alphabet}

\newcounter{minutes}
\divide\time by 60
\newcounter{hours}
\multiply\time by 60
\addtocounter{minutes}{-\time}

\renewcommand{\Re}{\mathop{\mathrm{Re}}}
\renewcommand{\Im}{\mathop{\mathrm{Im}}}
\renewcommand{\i}{\mathrm{i}}

\newcommand{\B}{{\mathbb B}}
\newcommand{\C}{{\mathbb C}}

\newcommand{\K}{{\mathcal K}}

\DeclareMathOperator{\sn}{\mathrm{sn}}
\DeclareMathOperator{\cn}{\mathrm{cn}}
\DeclareMathOperator{\dn}{\mathrm{dn}}

\DeclareMathOperator{\sign}{\mathrm{sign}}

\begin{document}

\bibliographystyle{amsplain}
\title[Image augmentation with conformal...]
{
Image augmentation
with conformal mappings 
for a convolutional neural network
}

\def\thefootnote{}
\footnotetext{
\texttt{\tiny File:~\jobname .tex,
          printed: \number\year-\number\month-\number\day,
          \thehours.\ifnum\theminutes<10{0}\fi\theminutes}
}
\makeatletter\def\thefootnote{\@arabic\c@footnote}\makeatother

\author[O. Rainio]{Oona Rainio}
\author[M.M.S. Nasser]{Mohamed M.S. Nasser}
\author[M. Vuorinen]{Matti Vuorinen}
\author[R. Kl\'en]{Riku Kl\'en}

\begin{abstract}
For augmentation of the square-shaped image data of a convolutional neural network (CNN), we introduce a new method, in which the original images are mapped onto a disk with a conformal mapping, rotated around the center of this disk and mapped under such a M\"obius transformation that preserves the disk, and then mapped back onto their original square shape. This process does not result the loss of information caused by removing areas from near the edges of the original images unlike the typical transformations used in the data augmentation for a CNN. We offer here the formulas of all the mappings needed together with detailed instructions how to write a code for transforming the images. The new method is also tested with simulated data and, according the results, using this method to augment the training data of 10 images into 40 images decreases the amount of the error in the predictions by a CNN for a test set of 160 images in a statistically significant way (p-value=0.0360).
\end{abstract}
\maketitle

\noindent\textbf{Author information.}\\ 
Oona Rainio$^1$, email: \texttt{ormrai@utu.fi},
ORCID: 0000-0002-7775-7656\\
Mohamed M.S. Nasser$^2$, email: \texttt{mms.nasser@wichita.edu}, ORCID: 0000-0002-2561-0978\\
Matti Vuorinen$^3$, email: \texttt{vuorinen@utu.fi}, ORCID: 0000-0002-1734-8228\\
Riku Kl\'en$^1$, email: \texttt{riku.klen@utu.fi}, ORCID: 0000-0002-0982-8360\\
1: Turku PET Centre, University of Turku and Turku University Hospital, Turku, Finland\\
2: Department of Mathematics, Statistics, and Physics, Wichita State University, Wichita, KS 67260-0033, USA\\
3: Department of Mathematics and Statistics, University of Turku, Turku, Finland\\

\section{Introduction}

A convolutional neural network (CNN) is a type of deep learning technique that is well-suited for processing images. It receives image data in a matrix format so that each element of the matrix corresponds to the value of one pixel in the image and then transforms this input through several layers by taking into account the spatial relationships between the data points. The CNNs have been noted to be very useful in different areas of research but training even a single CNN often requires a large number of labelled images, which can sometimes be difficult to obtain.

One possible solution to this problem is using data augmentation, which means that the amount of existing data is multiplied by using simple transformations to create new, slightly different versions of images. Typical transformations used for this purpose are rotations, reflections, and translations but they do not suit for all types of data. Namely, if the image is square-shaped, creating a new version of it with a translation always crops out some areas from the image close to its edges and there are only seven new images that can be created with such a rotation or a reflection that fully preserves the original square. Clearly, the issue would not be encountered if the images given to the CNN would be disk-shaped but this is rarely the case.  

However, according to the Riemann mapping theorem from classical function theory, any simply connected proper subdomain of the complex plane can be mapped onto the unit disk with a conformal mapping and, in particular, a conformal mapping called a Schwarz-Christoffel mapping can be used to map a two-dimensional disk onto the interior of a simple polygon \cite{dt}. Conformal mappings are much studied in complex analysis because they have many desirable properties such as preserving the magnitudes and directions of angles between curves even though they can turn straight line segments into circular arcs and vice versa. Furthermore, another subtype of conformal mappings are M\"obius transformations which can be defined so that they map the interior of a disk onto itself but still significantly transform its contents. While conformal mappings could potentially be utilized in image data augmentation, their formulas are generally quite complicated and require integration of complex valued functions that cannot be computed directly with the existing functions in common programming languages.

In this article, we study if image augmentation can be performed by first mapping each square-shaped image onto a disk with a conformal mapping, then applying such rotations and M\"obius transformations that preserve this disk, and finally mapping the resulting images back onto their original square shape. First, in Section 2, we present the usual formulas of conformal mappings and other mathematical theory needed. In Section 3, we show how these mappings can be written in Python or other programming languages and how images can be mapped with them. In Section 4, we test this method with one simple example about a CNN predicting a simulated data set. All the codes written in Python and MATLAB are also publicly available so the readers can access to these codes.

\section{Preliminaries}

Let $G$ be the interior of the square with the vertices $1+\i,-1+\i,-1-\i,1-\i$. In other words, define $G=\{z\in\C\text{ }|\text{ }-1<{\rm Re}(z)<1,-1<{\rm Im}(z)<1\}$, where $\C$ is the complex plane. Denote the unit disk $\{z\in\C\text{ }|\text{ }|z|<1\}$ by $\B^2$.

Consider the complete elliptic integrals of the first kind $\K(r)$ and $\K'(r)$ defined for $r\in(0,1)$ by~\cite{avv,nist}
\begin{equation}\label{eq:ellipK}
\K(r)=\int^1_0 \frac{dt}{\sqrt{(1-t^2)(1-r^2t^2)}},
\quad \K'(r)=\K(r'), \quad r'=\sqrt{1-r^2}.
\end{equation}
In many references, the complete elliptic integrals of the first kind $\K(r)$ are defined as in~\eqref{eq:ellipK}, see e.g. \cite{bateman,dali,kyth}. 
However, in this paper, we will use the notations used in~\cite{abra}. Thus, the complete elliptic integrals of the first kind are defined by~\cite[p.~590]{abra} 
\begin{equation}\label{eq:ellipKm}
	\K(m)=\int^1_0 \frac{dt}{\sqrt{(1-t^2)(1-mt^2)}},
	\quad \K'(m)=\K(m_1), 
\end{equation}
where $m=r^2\in(0,1)$ and $m_1=1-m$. 
The incomplete elliptic integral of the first kind is defined by~\cite{abra}
\begin{equation} \label{eq:ellipF}
F(\varphi,m)
=\int^{\varphi}_0 \frac{d\theta}{\sqrt{1-m \sin^2 \theta}}
=\int^{\sin \varphi}_{0} \frac{dt}{\sqrt{(1-t^2)(1-mt^2)}},
\end{equation}
and then $\K(r)=F(\pi/2,m)$. See \cite{abra,avv,bateman,dali,kyth} for more information. 

The Jacobian elliptic functions can be defined with the help of the incomplete elliptic integral~\eqref{eq:ellipF}, see~\cite[Ch.~16]{abra}.
If
\begin{align*}
	\phi=\int^{\varphi}_0 \frac{d\theta}{\sqrt{1-m\sin^2 \theta}},
\end{align*}
then the angle $\varphi$ is called the \emph{Jacobi amplitude} and we write
\[
\varphi = {\rm am}(\phi,m).
\]
The Jacobian elliptic functions are defined by
\begin{align}\label{eq:sncndn}
\sn(\phi,m)=\sin\varphi,\quad
\cn(\phi,m)=\cos\varphi,\quad
\dn(\phi,m)=\sqrt{1-m\sn^2(\phi,m)}.
\end{align}

The exact conformal mapping from the square region $G$ and its inverse can be written in terms of the elliptic functions and the incomplete elliptic integral. It follows from~\cite[p.~182]{kob} and~\cite[p.~242]{kyth} that the conformal mapping $f$ from the domain $G$ onto the unit disk $\B^2$, $f:G\to\B^2$, is given by 
\begin{align}\label{def_f}
	f(z)=\frac{1}{\sqrt{2}\,e^{\i\pi/4}}\frac{\sn(\sqrt{2}\,L\,z,m)}{\dn(\sqrt{2}\,L\,z,m)},\quad z\in G,
\end{align}
where 
\begin{align}\label{eq:rmL}
	r=\frac{1}{\sqrt{2}},\quad m=r^2=\frac{1}{2},\quad L=\frac{1}{2}\K(m)e^{\pi\i/4}.
\end{align}
Note that $\K(m)=\K(1/2)=\Gamma(1/4)^2/(4\pi)$, see \cite[(3.19), p.~51]{avv}.
Then, it follows from~\eqref{eq:sncndn} and~\eqref{def_f} that the inverse of this mapping, $f^{-1}:\B^2\to G$, is given
\begin{align}\label{def_invf}
	f^{-1}(z)=\frac{1}{\sqrt{2}L}F\left(\frac{\sqrt{2}e^{\pi\i/4}z}{\sqrt{1+z^2\i}},m\right),\quad z\in\B^2,    
\end{align}
which is a conformal mapping from the unit disk $\B^2$ onto the square domain $G$.

Denote the extended complex space by  $\overline{\C}=\C\cup\{\infty\}$ and choose some $\alpha\in\B^2$. Define the M\"obius transformation $g:\overline{\C}\to\overline{\C}$,
\begin{align}\label{def_mob}
g(z)=\frac{z-\alpha}{1-\overline{\alpha}z},\quad z\in\C,    
\end{align}
where $\overline{\alpha}$ is the complex conjugate of $\alpha$. This mapping $g$ fulfills $g(\B^2)=\B^2$, $g(\alpha)=0$, and $g(0)=-\alpha$, which means that it preserves the unit disk $\B^2$ but is not an identity mapping as long as $\alpha\neq0$.

Finally, we define formally the rotation about the origin for an angle $k\in[0,2\pi)$ as $\upsilon:\C\to\C$,
\begin{align}\label{def_rot}
\upsilon(z)=ze^{ki},\quad z\in\C.   
\end{align}

\begin{figure}[!tbp]
  \centering
  \begin{subfigure}[b]{0.35\textwidth}
  \centering
  \includegraphics[width=\textwidth]{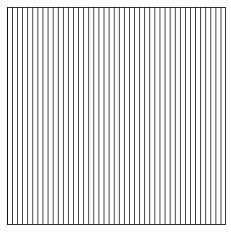}
  \caption{Original line image}
  \end{subfigure}
  \hspace{1cm}
  \begin{subfigure}[b]{0.35\textwidth}
  \centering
  \includegraphics[width=\textwidth]{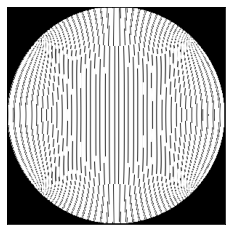}
  \caption{Image after the mapping $f$}
  \end{subfigure}
  \\
  \begin{subfigure}[b]{0.35\textwidth}
  \centering
  \includegraphics[width=\textwidth]{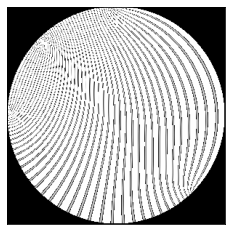}
  \caption{Image after $g\circ f$}
  \end{subfigure}
  \hspace{1cm}
  \begin{subfigure}[b]{0.35\textwidth}
  \centering
  \includegraphics[width=\textwidth]{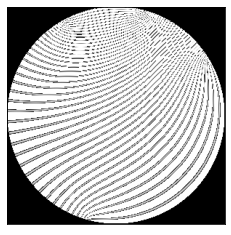}
  \caption{Image after $\upsilon\circ g\circ f$}
  \end{subfigure}
  \\
  \begin{subfigure}[b]{0.35\textwidth}
  \centering
  \includegraphics[width=\textwidth]{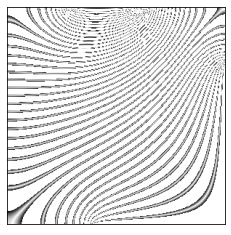}
  \caption{Image after $f^{-1}\circ\upsilon\circ g\circ f$}
  \end{subfigure}
  \hspace{1cm}
  \begin{subfigure}[b]{0.35\textwidth}
  \end{subfigure}
  \caption{An image of parallel line segments before and after different mappings, including the conformal mapping $f$ from the original square onto the unit disk, the M\"obius transformation $g$ defined with $\alpha=0.3+0.3\i$, a rotation $\upsilon$ with $k=\pi/3$, and the inverse mapping $f^{-1}$.}
  \label{fig1}
\end{figure}

Figure \ref{fig1} represents visually how the functions $f$, $g$, and $\upsilon$ and the inverse function $f^{-1}$ affect the shape of an image, when $g$ and $\upsilon$ are defined with constants $\alpha=0.3+0.3\i$ and $k=\pi/3$, respectively.

\section{Code}

Computing the mapping function $f$ by~\eqref{def_f} and its inverse function $f^{-1}$ by~\eqref{def_invf} requires computing the Jacobian elliptic functions $\sn(\phi,m)$ and $\dn(\phi,m)$ for complex $\phi$ as well as computing the incomplete elliptic integral $F(\varphi,m)$  for complex $\varphi$. Thus, the function $f$ and its inverse function $f^{-1}$ cannot usually be computed directly from their definitions \eqref{def_f} and \eqref{def_invf} in the code because the functions used to evaluate the elliptic integrals are not defined for complex numbers in many programming languages. 
For instance, the complete elliptic integral $\K(m)$, the incomplete elliptic integral $F(\varphi,m)$, as well as the Jacobian elliptic functions $\sn(\phi,m)$, $\cn(\phi,m)$, and $\dn(\phi,m)$ can be computed for real arguments, , with the functions \texttt{ellipk}, \texttt{ellipkinc}, and \texttt{ellipj} of the subpackage \texttt{special} of the package Scipy \cite{v20} in Python.
So, in this paper, we will use the properties of the Jacobian elliptic functions and the incomplete elliptic integral to compute their values without the need to compute elliptic integrals of complex variables. 
For example, the functions $\sn(z,m)$ and $\dn(z,m)$, for a complex variable $z=x+\i y$ where $x$ and $y$ are real variables, can be computed using the following formulas~\cite[\S16.21.1]{abra}
\begin{eqnarray}
	\sn(z,m)&=&\frac{s_0d_1+\i\,c_0d_0s_1c_1}{\delta},\\
	\dn(z,m)&=&\frac{d_0c_1d_1-\i\,ms_0c_0s_1}{\delta},
\end{eqnarray}
where
\begin{eqnarray*}
&&s_0=\sn(x,m), \quad c_0=\cn(x,m), \quad d_0=\dn(x,m), \\ 
&&s_1=\sn(y,1-m), \quad c_1=\cn(y,1-m), \quad d_1=\dn(y,1-m),  
\end{eqnarray*}
and
\[
\delta=c_1^2+ms_0^2s_1^2.
\]

Similarly, for a complex variable $z=x+\i y$, the incomplete elliptic integral $F(z,m)$ can be computed by~\cite[\S17.4.11]{abra}
\begin{equation}
	F(z,m)=F(x+\i y,m)=F(x_1,m)+\i\,F(y_1,m_1),
\end{equation}
where $x_1$ and $y_1$ are real variables such that $X=\cot^2 x_1$ is the positive root of the equation
\begin{align}\label{equx}
	X^2-\left(\cot^2(x)+m\sinh^2(y)\csc^2(x)-m_1\right)X - m_1\cot^2(x)=0,
\end{align}
and $\tan^2y_1$ is given by
\begin{align}\label{equy}
	\tan^2y_1=\left(\tan^2x\cot^2(x_1)-1\right)/m.   
\end{align}
Note that, if $b=-\left(\cot^2(x)+m\sinh^2(y)\csc^2(x)-m_1\right)$ and $c=- m_1\cot^2(x)$, then $c\le0$ and hence $\sqrt{b^2/4-c}\ge b/2$. Thus, we have
\begin{align}\label{equx1}
	\cot^2x_1=-b/2+\sqrt{b^2/4-c}.   
\end{align}
Denote below the floor and the ceiling functions with ${\rm floor}()$ and ${\rm ceil}()$.
Then, the values of $x_1$ and $y_1$ can be computed from~\eqref{equx1} and~\eqref{equy}, respectively, through (see~\cite{mois})
\begin{eqnarray}
	x_1&=&\sign(\cot(x))\cot^{-1}\sqrt{-b/2+\sqrt{b^2/4-c}}+\pi {\rm ceil}(x/\pi-0.5),\\
	y_1&=&\sign(y)\tan^{-1}\sqrt{\left(\tan^2x\cot^2(x_1)-1\right)/m}.
\end{eqnarray}

At the beginning of the code, fix the constants $r$, $m$, and $L$ as in~\eqref{eq:rmL}. The conformal mapping $f$ from the square $G$ onto the unit disk can now be written with the following pseudo-code.


\begin{equation}\label{cod_f}
\begin{aligned}
{\rm define}\quad &f(z):\\
&\#\text{input: a complex number $z$ with Re($z$),Im($z$)$\in$[-1,1]}\\
&\#\text{output: a complex number $w$ with $|w|<1$}\\
&\hat z=\sqrt{2}Lz\\
&x=\Re\hat z;\quad y=\Im\hat z\\
&s_0=\sn(x,m), \quad c_0=\cn(x,m), \quad d_0=\dn(x,m), \\ 
&s_1=\sn(y,1-m), \quad c_1=\cn(y,1-m), \quad d_1=\dn(y,1-m),  \\
&\delta=c_1^2+ms_0^2s_1^2\\
&{\rm sni}=(s_0d_1+\i\,c_0d_0s_1c_1)/\delta\\
&{\rm dni}=(d_0c_1d_1-\i\,ms_0c_0s_1)/\delta\\
&w=\frac{\rm sni}{\sqrt{2}\,e^{\pi\i/4}\,\rm dni}\\
&{\rm return}(w)    
\end{aligned}    
\end{equation}

Finding the inverse mapping $f^{-1}$ requires solving the quadratic equation~\eqref{equx}.
The more unusual trigonometric functions such cot or csc and their inverses are in the package \texttt{sumpy} in Python. Note also that the parameter $\epsilon$ can be chosen to be any small positive number.

\begin{equation}\label{cod_invf}
\begin{aligned}
{\rm define}\quad &f^{-1}(z):\\
&\#\text{input: a complex number $z$ with $|z|<1$}\\
&\#\text{output: a complex number $w$ with Re($w$),Im($w$)$\in$[-1,1]}\\
&\hat z=\frac{\sqrt{2}e^{\pi\i/4}z}{\sqrt{1+z^2\i}}\\
&\tilde z=\arcsin(\hat z)\\
&x=\Re \tilde z;\quad y=\Im \tilde z;\quad\epsilon=0.00001\\
&\text{{\rm \#To avoid singularity of cot($x$) at zero add }}\epsilon\\
&{\rm if }|x|<\epsilon:\\
&\quad\quad\,x=\epsilon\\
&\text{{\rm \#Find the roots of the equation \eqref{equx}}}\\
&b=-\left(\cot^2(x)+m\sinh^2(y)\csc^2(x)-m_1\right);\quad c=- m_1\cot^2(x)\\
&X_1=-b/2+\sqrt{b^2/4-c}\\
&x_1={\rm arccot}(\sqrt{X_1})\\
&{\rm if }\, X_1\tan^2(x)<1:\\
&\quad\quad\,y_1=0\\
&{\rm else}:\\
&\quad\quad\,y_1=\arctan\left(\sqrt{\left(X_1\tan^2(x)-1\right)/m}\right)\\
&\text{{\rm \#Change of variables taking into account the periodicity ceil to the right}}\\
&x_1=(-1){\rm floor}(2x/\pi)x_1+\pi{\rm ceil}(x/\pi-0.5+\epsilon);
\quad y_1={\rm sign}(y)y_1\\
&F_1=F(x_1,m);\quad F_2=F(y_1,1-m)\\
&w=(F_1+\i\,F_2)/(\sqrt{2}L)\\
&{\rm return}(w)
\end{aligned}    
\end{equation}

After writing the functions $f$ and $f^{-1}$ with the pseudo-codes \eqref{cod_f} and \eqref{cod_invf}, they can be tested by choosing a random number $z$ from the square $G$, computing $w=f(z)$ and $z'=f^{-1}(w)$, and printing the difference $z-z'$, which should be very close to zero.

An image can be mapped conformally onto a disk with the function squareToDisk presented in \eqref{func_std}. First, suppose then that there is a square matrix called img which can be read as a square-shaped grayscale image by choosing the colour of each pixel according the corresponding element in img. Let $h$ be the number of rows or columns in the matrix img, imgH a vector containing $h$ evenly spaced points in the interval $[-1,1]$, and $u$ the distance between two adjacent points in imgH. Then we initialize the new image img by creating a zero matrix of the same size as img. After that, we create a loop that goes through each element of img and expresses it as a point $z\in G$ with the help of the vector imgH. If the point $z$ belongs to the unit disk, we use the inverse mapping $f^{-1}$ to find the point $w$ in the square $G$ that becomes $z$ when the domain $G$ is mapped conformally to the unit disk with the mapping $f$. For this point $w$, we find such points $j_0,k_0\in(0,1,...,h-1)$ and $j_1,k_1\in[0,u)$ that $w=j_0+j_1+(k_0+k_1)i$, which gives us also the closest four points to $w$ that correspond to the pixel locations of the original image matrix img and the distances between $w$ and these locations. Finally, we compute the values of the pixel at the point $w$ by using the weighted means of the four surrounding pixels and, since $z=f(w)$, we have the value of the pixel at the point $z$ in the unit disk. In the pseudo-code below, $v[i]$ refers to the $(i+1)$th element of the vector $v$ as the indexing starts from 0.  

\begin{equation}\label{func_std}
\begin{aligned}
{\rm define}\quad &{\rm squareToDisk}({\rm img}):\\
&\#\text{input: a square-matrix img presenting some image}\\
&\#\text{output: a square-matrix img1 containing the transformed image like Fig. 1(B)}\\
&h={\rm dim}({\rm img})[0]\\
&{\rm imgH}=\frac{2}{h-1}(0,1,...,h-1)-(1,...,1)\\
&u={\rm imgH}[1]-{\rm imgH}[0]\\
&{\rm img1}=0_{{\rm dim}({\rm img})}\\
&\text{{\rm for }} j\text{{\rm{ }in }}(0,1,...,h-1):\\
&\quad\quad\,\text{{\rm for }} k\text{{\rm{ }in }}(0,1,...,h-1):\\
&\quad\quad\quad\quad\,\,z={\rm imgH}[j]+{\rm imgH}[k]\i\\
&\quad\quad\quad\quad\,\,\text{{\rm if }}|z|<1:\\
&\quad\quad\quad\quad\quad\quad\,\,\,w=f^{-1}(z)\\
&\quad\quad\quad\quad\quad\quad\,\,\,j_0={\rm floor}(({\rm Re}(w)+1)/u)\\
&\quad\quad\quad\quad\quad\quad\,\,\,j_1=({\rm Re}(w)+1)/u-j_0\\
&\quad\quad\quad\quad\quad\quad\,\,\,k_0={\rm floor}(({\rm Im}(w)+1)/u)\\
&\quad\quad\quad\quad\quad\quad\,\,\,k_1=({\rm Im}(w)+1)/u-k_0\\
&\quad\quad\quad\quad\quad\quad\,\,\,{\rm img1}[j,k]=j_1k_1{\rm img}[j_0,k_0]+(1-j_1)k_1{\rm img}[j_0+1,k_0]+\\
&\quad\quad\quad\quad\quad\quad\quad\quad\quad\quad\quad\,\,\,\,\,j_1(1-k_1){\rm img}[j_0,k_0+1]+\\
&\quad\quad\quad\quad\quad\quad\quad\quad\quad\quad\quad\,\,\,\,\,(1-j_1)(1-k_1){\rm img}[j_0+1,k_0+1]\\
&{\rm return(img1)} 
\end{aligned}    
\end{equation}

The function in \eqref{func_std} returns such an image matrix that has the original image mapped onto the largest possible disk fitting inside the square-shaped image matrix, and the other values outside this disk are zeroes. Note that this function can be extended for also colour images of RGB or another similar format by just computing the weighted means for the values of each color channel at the end of the loop. Similarly, by replacing the inverse function $f^{-1}$ with either the M\"obius transformation $g$ of \eqref{def_mob} or the rotation $\upsilon$ of \eqref{def_rot}, we can map the input image so that the part inside the disk is transformed. To create the function that conformally maps the interior of this disk in the square-shaped image matrix onto the whole matrix, we just need to remove the condition $|z|<1$ and replace $w=f^{-1}(z)$ by $w=f(z)$ in the code \eqref{func_std}. Alternatively, we can use for instance the composed mapping $w=(f^{-1}\circ\upsilon\circ g\circ f)^{-1}(z)=f^{-1}\circ g\circ\upsilon\circ f(z)$ to obtain the image of Figure \ref{fig1}(E).

\section{Experiment}

Here, we build a CNN for predicting how many small black disks an otherwise white image contains. We use Python (version: 3.9.9) \cite{pyt09} with packages TensorFlow (version: 2.7.0) \cite{tf15}, Keras (version: 2.7.0) \cite{k15}, and SciPy \cite{v20} (version: 1.7.3).

\begin{figure}[!tbp]
  \centering
  \begin{subfigure}[b]{0.23\textwidth}
  \centering
  \includegraphics[width=\textwidth]{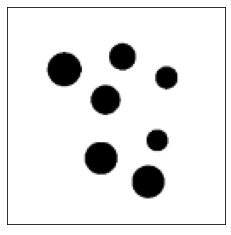}
  \caption{Original image\\
  \color{white}{.}\\
  \color{white}{.}}
  \end{subfigure}
  \hspace{0.5cm}
  \begin{subfigure}[b]{0.23\textwidth}
  \centering
  \includegraphics[width=\textwidth]{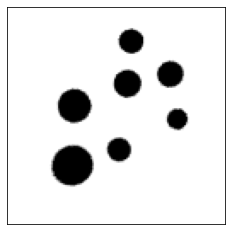}
  \caption{Augmented version with $\alpha=0.1+0.1\i$ and $k=\pi/3$}
  \end{subfigure}
  \\
  \begin{subfigure}[b]{0.23\textwidth}
  \centering
  \includegraphics[width=\textwidth]{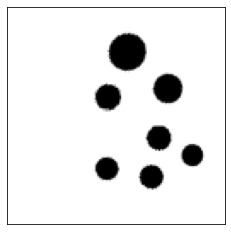}
  \caption{Augmented version with $\alpha=0.1+0.3\i$ and $k=\pi$}
  \end{subfigure}
  \hspace{0.5cm}
  \begin{subfigure}[b]{0.23\textwidth}
  \centering
  \includegraphics[width=\textwidth]{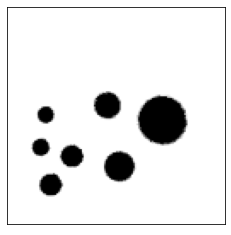}
  \caption{Augmented version with $\alpha=0.3+0.3\i$ and $k=3\pi/2$}
  \end{subfigure}
  \caption{An image with $n=7$ disks and three augmented versions of this image that have been created with the composed mapping $f^{-1}\circ\upsilon\circ g\circ f$, where the mappings $f,g,\upsilon$ are as \eqref{def_f}-\eqref{def_rot} for the specified choices of $\alpha$ and $k$.}
  \label{fig3}
\end{figure}

\subsection{Data}

The data set consisted of a collection of images from one to ten black disks on white background and an explaining variable containing the correct number of disks for each image. For a number $n=1,...,10$, a single image was created by first initializing a $300\times300$ null matrix corresponding to the domain $G$, choosing $n$ centers $c_j\in[-0.7,0.7]\times[-0.7,0.7]$ such that $\min_{j,k\in\{1,...,10\}}\{|c_j-c_k|\}>0.4$ and $n$ radii $R_j\in[0.1,0.17]$, and changing each element of the matrix from 0 to 1 if and only if the distance $c_j$ was less than $R_j$ for some $j$. Out of 170 different images, 10 were included into the training set and the rest 160 into the test set. Augmented version of the training set with 40 images was then created by adding three versions of each image in the original training set by mapping them with the composed mapping $f^{-1}\circ\upsilon\circ g\circ f$, where $f,g,\upsilon$ are as in \eqref{def_f}-\eqref{def_rot} for $\alpha=0.1+0.1\i$ and $k=\pi/3$, $\alpha=0.1+0.3\i$ and $k=\pi$, and $\alpha=0.3+0.3\i$ and $k=3\pi/2$, as shown in Figure \ref{fig3}. We also created another augmented training set of 40 images by using rotations of $k$ degrees, $k\in(-15,15)$. The final images were scaled to the size of $128\times128$ pixels.

\subsection{Convolutional neural network}

The CNN used here is the same as in \cite{h23}. It is based on the U-Net architecture introduced in \cite{r15}, which is commonly used in segmentation of medical images. While a typical U-Net contains first an expanding path and then a contracting path to perform the segmentation, our CNN only has the contracting path after which it returns a single numeric value. The contracting path consist of four sequences, each of which contains two convolutions and one pooling operation, and, after that, there are four dense layers. The activation function of the last layer is a linear function and, for all the other layers, the ReLu function is used. The CNN was trained on 130 epochs for the non-augmented data set and two augmented data sets by using Adam as the optimizer, the mean squared error (MSE) as the loss function with learning rate of 0.001, and validation split of 30\%.

\subsection{Methods}

The CNN is first initialized, trained with the non-augmented data set, and used to predict the values of the test set. Then we compute the squared error between the predicted number of disks and the real number of disks for each image of the test set. After that, the CNN is re-initialized to its initial state and the experiment is re-run by using the augmented training data sets instead. The three methods are compared by computing the MSE as the mean value of the squared errors and, to see if the differences in these means are statistically significant or not, the Student's t-test is performed for the distributions of the squared errors.

\subsection{Results}

The MSE of the predictions of the test set was 1.742 when the CNN was trained for the data augmented by using conformal mappings. The corresponding MSE was 2.381 for non-augmented data and 2.095 for the data augmented with just rotations. According the Student's t-test between the squared errors of the predictions from the non-augmented and the augmented model, the difference between the MSEs was statistically significant with a p-value 0.0360. However, the difference between two different augmentation types was not statistically significant (p-value=0.196). Pearson's correlation coefficient was 0.884 between the predictions on the test set by the non-augmented CNN and the real numbers of disks, 0.883 for the rotation augmentation, and 0.924 for the conformal mapping augmentation (Fig. \ref{fig3}). 

\begin{figure}
    \centering
    \includegraphics[scale=0.7]{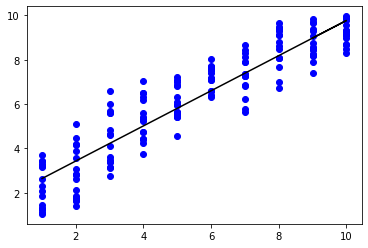}
    \caption{The predictions of the augmented CNN against the real values of the test set with the least squares regression line. The slope of the line is 0.793 and the intercept is 1.846. The correlation between the predictions and the real values is 0.924.}
    \label{fig4}
\end{figure}

\section{Discussion}

Firstly, it must be noted that the experiment above is very simplified example and better results could be obtained by building a suitable algorithm instead of using a CNN. However, there are numerous practical reasons why a CNN that recognizes certain unusual disk-shaped areas are useful. In fact, this experiment was inspired by the CNNs detecting tumors from tomography images of cancer patients in \cite{h23,l23,lim}.

It also should be taken into account that this method is quite complicated and designed for situations where the typical transformations cannot be used. For instance, if some of the disks would be so close to edges of the images in our experiment that they would be cropped completely or partially out when the square-shaped images are directly rotated and re-calculating their number is not possible, using this new method rather than the usual rotation is justified. The rotation used as a comparison augmentation method in our experiment did not crop the images but rather included more background. Given how commonly CNNs are used nowadays, there are likely many practical examples of cases where information could be lost if the typical rotation is used. Note that these mappings can be also extended to map any rectangle-shaped image into a disk or vice versa because the rectangle can be very easily stretched into a square.

Another reason for using this method is wanting to utilize the properties of the M\"obius transformation in particular. Because of this mapping, the images change in more complicated ways under this augmentation method than they do in simple rotations. We can see that the right-most disk in Figure \ref{fig3}(D) is larger than any of the disks in the original image in Figure \ref{fig3}(A). By increasing the absolute value of the point $\alpha$ used to define the M\"obius transformation, the differences between the images before and after the composed mapping are also increased. Still, all these disks stay circular because M\"obius transformations can only map circles into lines or circles and the disks stay inside the edges of the image. The use of M\"obius transformations in image augmentation has been also studied by Zhou et al. \cite{z21} but they did not use other conformal mappings and their augmented images therefore contain much empty background outside the original photographs.

One alternative method is using general adversarial network (GAN) augmentation first introduced in 2014 by Goodfellow et al. \cite{g14}. GANs are a class of neural networks that generate synthetic samples resembling the real images of the original data set. However, it might be difficult to predict what sort of augmented data GANs produce while the use of conformal mappings only distort the images. This means that the augmentation based on the conformal mappings preserves the number of disks in the images of the data set used here, while GANs might change the number of disks. Furthermore, GANs also require some amount of data so that they can be trained for their work, while the amount of the data does not affect how the images change under the conformal mappings. Our method of augmentation could be also compared with such procedures that use prior information about the data distribution in augmentation \cite{b22,n98}.

\section{Conclusion}

We used conformal mappings to create a new way to augment the image data of a CNN and, according to our result, this method both works and produces better predictions than a CNN trained with non-augmented data set.\\

\noindent\textbf{Data and code availability statement.} Available at \url{https://github.com/rklen/Conf_map_augmentation}\\
\textbf{Conflict of interest statement.} On the behalf of all authors, the corresponding author states that there is no conflict of interest.\\
\textbf{Funding.} The first author was financially supported by the Finnish Culture Foundation and Magnus Ehrnrooth Foundation.\\
\textbf{Acknowledgments.} The authors are grateful to the referees for their valuable comments.

\end{document}